\documentclass[a4paper]{./styles/svproc}

\usepackage{graphicx}
\usepackage{url}

\begin{document}
\mainmatter              % start of a contribution

% We welcome submissions of up to 16 pages (including references)
%
\title{Self-Deployable, Adaptive Soft Robots Based on Contracting-Cord Particle Jamming}
\titlerunning{Self-Deployable, Adaptive Soft Robots}  % abbreviated title (for running head)
\author{Wenzhong Yan\inst{1,2\ast} \and Brian Ye\inst{1}
\and Mingxi Li\inst{2} \and \\ Jonathan B. Hopkins\inst{1} \and Ankur Mehta\inst{2}}
\authorrunning{Wenzhong Yan et al.} % abbreviated author list (for running head)

\institute{Mechanical and Aerospace Engineering Department, UCLA, USA\\
\email{$^\ast$ wzyan24@g.ucla.edu} \and
Electrical and Computer Engineering Department, UCLA, USA}

\maketitle            % typeset the title of the contribution

\begin{abstract}
We developed a new class of soft locomotive robots that can self-assemble into a preprogrammed configuration and vary their stiffness afterward in a highly integrated, compact body using contracting-cord particle jamming (CCPJ). We demonstrate this with a tripod-shaped robot, TripodBot, consisting of three CCPJ-based legs attached to a central body. TripodBot is intrinsically soft and can be stored and transported in a compact configuration. On site, it can self-deploy and crawl in a slip–stick manner through the shape morphing of its legs; a simplified analytical model accurately captures the speed. The robot's adaptability is demonstrated by its ability to navigate tunnels as narrow as 61\% of its deployed body width and ceilings as low as 31\% of its freestanding height. Additionally, it can climb slopes up to 15 degrees, carry a load of 5 grams (2.4 times its weight), and bear a load 9429 times its weight.

\end{abstract}

\begin{figure*}[h]
  \centering
  \includegraphics[trim=0.6in 11.2cm 0.7in 0in, clip=true, width=1\textwidth]{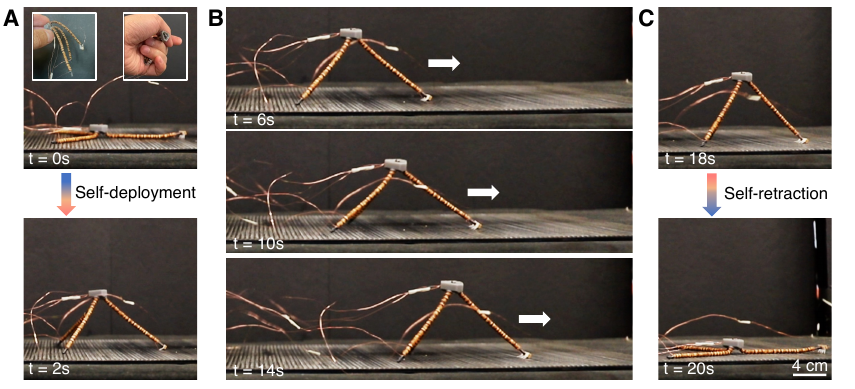}
  \caption{\textbf{The TripodBot in action.} (A) The TripodBot can self-assemble from a soft, compact state to a preprogrammed tripod shape with certain rigidity. (B) After deployment, it can crawl in a slip-slick manner. (C) After finishing the task, it can collapse back to soft state, ready for the next deployment.}
  \label{fig:teaser}
\end{figure*}

\section{Introduction}
Locomotive robots capable of self-assembly and post-deployment stiffness and shape variance are of great interest\cite{chen2024scale}. These robots hold vast potential for applications in exploration within complex and remote areas, where they can be transported in compact conformations and deployed into preprogrammed configurations for meaningful onsite tasks\cite{felton2014method}. Due to their tunable shape and stiffness\cite{yan2024self}, such robots can adapt to complex environments, such as exploring narrow gaps in rubble \cite{lathrop2023directionally}. However, creating such self-deployable, adaptive robots remains a significant challenge, notwithstanding its importance.

Recently, there has been increasing interest towards achieving this goal\cite{jayaram2016cockroaches,sun2023embedded,stilli2014shrinkable,babu2019antagonistic}. Drawing inspiration from biological organisms, soft robots are designed to achieve high flexibility and environmental adaptability due to their composition of primarily compliant materials\cite{laschi2016soft}. To enhance the load-carrying capacity and structural rigidity of soft robots, various variable-stiffness mechanisms have been proposed, including low-melting-point alloys \cite{hwang2022shape},  magnetorheological fluids \cite{majidi2010tunable}, and antagonistic actuation \cite{althoefer2018antagonistic}. These methods often require the integration of multiple actuators with coordinated control \cite{althoefer2018antagonistic} or the addition of passive variable-stiffness modules alongside actuation modules for shape change \cite{yang2021reprogrammable}. Such integrations lead to bulky robotic systems \cite{wall2015selective} and complex control architectures \cite{narang2018mechanically}. Although these approaches offer greater system flexibility, allowing independent control of robots' stiffness and shape, they are challenging to incorporate into locomotive robots, especially at the centimeter scale. Therefore, there is a strong need to develop self-deployable, adaptive soft robots utilizing a single actuation system to minimize implementation complexity and weight.

Here, we introduce a new class of soft robots that can self-deploy and adjust both shape and stiffness post-assembly in a highly integrated manner. These robots build upon our previous work on contracting-cord particle jamming (CCPJ) \cite{yan2024self}. The CCPJ mechanism primarily involves contracting actuators threaded through beads with matching interfaces in networked chains with initial slack. When the contracting actuators are activated, the beads gather into preprogrammed configurations, and further contraction tunes the stiffness of the assemblies, similar to common material jamming \cite{hasegawa2023tension,amend2012positive}. Compared to vacuum-driven active particle jamming \cite{chen2024scale}, CCPJ offers several advantages: (1) it is easier to fabricate without requiring airtight sealing, and thus provides higher scalability; (2) it does not require bulky associated components, such as pumps and valves, for untethered operation; and (3) it is robust to punctures, which are difficult to avoid in complex, restricted areas.

We demonstrate the capabilities of this technology with a tripod-shaped robot, TripodBot, featuring three CCPJ-based legs attached to a central body (Fig.\ref{fig:teaser}). We harness shape memory alloys (SMA) as main contracting-cord actuators. TripodBot is intrinsically soft and can be stored and transported in a compact configuration. Once deployed, it crawls using slip-stick locomotion facilitated by the shape morphing of its legs. The locomotion speed of TripodBot varies with changes in the actuation period, reaching a maximum speed of approximately  0.51 m/min (7.5 BL/min), accurately predicted by a simplified analytical model. The robot's adaptability is showcased by its ability to navigate tunnels as narrow as 61\% of its deployed body width and ceilings as low as 31\% of its freestanding height. This adaptability is prominent among similar robots\cite{lathrop2023directionally}.
Additionally, TripodBot can climb slopes up to 15 degrees and carry a load of 5 grams (2.4 times its own weight). It can also bear a load, 9429 times its weight when incorporating super-coiled polymer (SCP) actuators\cite{yan2023towards} instead of SMA wires. This work offers a practical framework for designing soft robots with self-deployability, adaptability, and load-carrying capability, suitable for exploration in remote and complex areas.

Specifically, the contributions of this paper include:

\begin{enumerate}
    \item a locomotion mechanism through the combination of directional friction and shape-changing of CCPJ mechanism,
    \item a method that enables adaptive locomotion and load-carrying through CCPJ, and
    \item a soft crawling robot fabricated out of the proposed design and verified by experiments.

\end{enumerate}

The remainder of the paper is organized as follows: in Section \ref{sec:CCPJ}, we briefly introduce the CCPJ mechanism; in Section \ref{sec:Design&Fabrication}, we introduce the design, fabrication, and control of TripodBot; in Section \ref{sec:Demo}, we demonstrate the performance of TripodBot; and the conclusion is presented in Section \ref{sec:Conclusions}.

\section{CCPJ for Self-Deployment and Stiffness Variance}
\label{sec:CCPJ}
\begin{figure*}[ht]
  \centering
  \includegraphics[trim=0in 12.2cm 0in 0in, clip=true, width=1\textwidth]{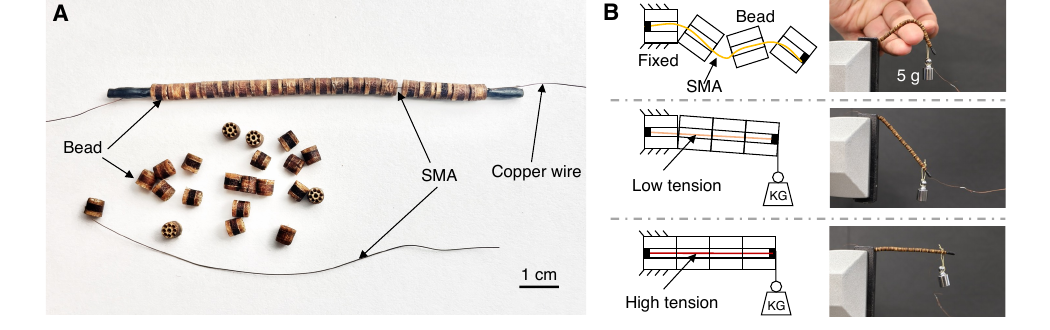}
  \caption{\textbf{Design and fabrication of a CCPJ beam.} (A) The beam is composed of 20 beads threaded through with a SMA wire. Each assembled beam weights 0.46 g. (B) After applying current across the SMA wire, the beam can self-deploy and continuously increases its stiffness as the current increase. One beam with a current of 0.4 A can hold a 5 g weight.}
  \label{fig:CCPJ}
\end{figure*}

CCPJ structures are the main components for constructing our self-deployable, adaptive robots. Therefore, we will introduce the fundamental mechanism and basic characteristics of CCPJ structures here, preparing the design of our robots. CCPJ structures are comprised of contracting actuators threaded through beads with matching interfaces in networked chains\cite{yan2024self}. The fundamental unit is a straight CCPJ beam as the leg of the TripodBot (Fig.\ref{fig:teaser}). Therefore, we use this straight beam as an example to show the characteristics of CCPJ structures. As shown in Fig.\ref{fig:CCPJ}, the slack network conforms to arbitrary shapes, but when actuated, it self-assembles into a preprogrammed configuration with beads gathered together. Further contraction of the actuators can dynamically tune the assembly's mechanical properties through the beads' particle jamming, while maintaining the overall structure with minimal change. 

\subsection{Design, Fabrication, and Experimental validation of CCPJ}
The CCPJ beam (Fig.\ref{fig:CCPJ}A) mainly consists of 20 beads (3 mm thickness, plywood) threaded through the central hole by a SMA wire of 0.15 mm diameter (Biometal). Two ends of the SMA wire are connected to an electrical power source through copper wires and fixed firmly to specify the length with heat shrinking tubes. Here we chose to use SMA given the potential for untethered operation with widely available electrical control and power\cite{huang2018chasing}. Other contracting actuator can be used for different goals. Each beam has an initial slack of 1.6 mm, which means the length of the SMA wire is 1.6 mm longer than the total length of the 20 beads. Therefore, the CCPJ beam is soft before actuation (Fig.\ref{fig:CCPJ}B). When current is applied to the SMA wire, the beads conform to the preprogrammed straight configuration with certain compliance. As the current increases, the beam becomes stiffer and can sustain a 5 gram weight with minimal deflection (about 11 times its own weight, Fig.\ref{fig:CCPJ}B).

\begin{figure*}[ht]
  \centering
  \includegraphics[trim=0in 11.6cm 0in 0in, clip=true, width=1\textwidth]{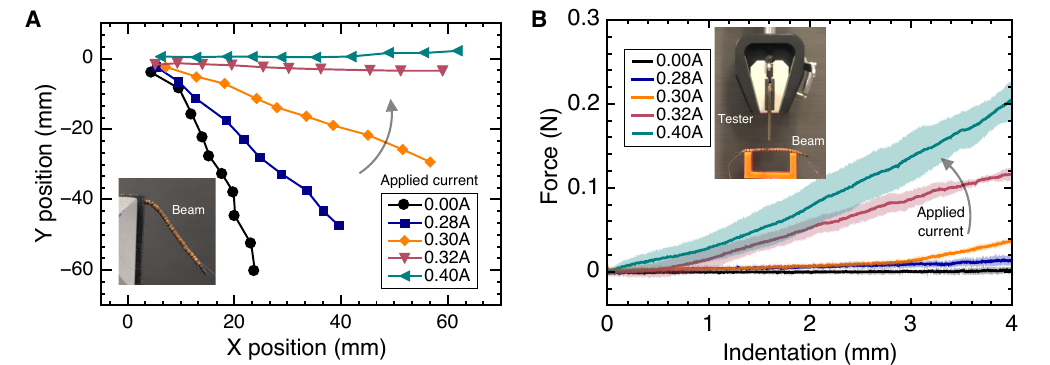}
  \caption{\textbf{Characterization of a CCPJ beam.} (A) The curvatures of the horizontally clamped beam against gravity with varying applied current on the SMA wire. (B) The force-indentation relationship changes over the applied current. Shaded areas represent SD over three tests.} 
  \label{fig:characterization}
\end{figure*}

\subsection{Self-Deployment of CCPJ Beam}
To quantitatively evaluate the self-deployment of the CCPJ beam, we clamped the beam horizontally against gravity(Fig.\ref{fig:characterization}A). We recorded the shapes of the CCPJ beam under various current ranging from 0 A to 0.4 A. Then we extracted the curvatures of the beam using \textit{Tracker} and plotted them in Fig.\ref{fig:characterization}A. When the current reached 0.32A, the beam became straight. Increasing current did not change the shape, however, the rigidity of the beam can be increased(see the next section). Here, we only consider the effect of gravity on the beam. Different loading conditions can change how the shape evolution along the change of the applied current.

\subsection{Stiffness Variation of CCPJ Beam}

To characterize the stiffness of the CCPJ beam under different currents, we employed a mechanical tester (Mecmesin). The indentor of the tester moved downward at a speed of 6 mm/min and measured the force acted upon the indentor in the vertical direction. The tested CCPJ beam was simply supported by two acrylic beams with a distance of 40 mm (Fig.\ref{fig:characterization}B). We measured the force-indentation curves of the CCPJ beam under nine different current levels. For each current level, we repeated the test three times to obtain the average force-indentation curves. The shaded areas in Fig.\ref{fig:characterization}B represents the SD. Between each test, we manually straightened the beam.  

\begin{figure*}[ht]
  \centering
  \includegraphics[trim=0in 12cm 3.7in 0in, clip=true, width=0.56\textwidth]{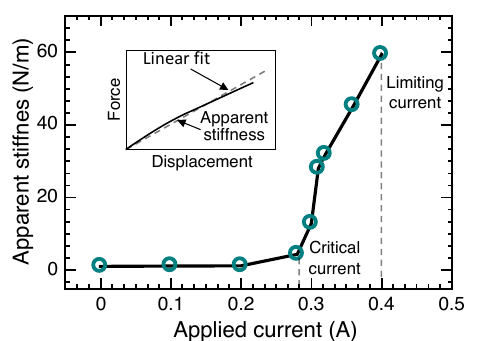}
  \caption{\textbf{Apparent stiffness variance over applied current of a CCPJ beam.} The stiffness can increase by 53.7 times when the current changes from 0 A to 0.4 A.}
  \label{fig:stiffness_current}
\end{figure*}

The results show that the stiffness of the beam increased monotonically with the increasing current. The elasticity of the whole beam was rather linear within our deflection range (0–4 mm). The apparent bending stiffness of the CCPJ beam under various applied currents was calculated and is summarized in Fig.\ref{fig:stiffness_current}. As the current increased from 0 to 0.4 A, the apparent bending stiffness of the structure exhibited
a rapid increase from 1.1 to 59.1 N/m with a 53.7 times increase. Note that, when the current exceeded 0.28 A, this apparent bending increase became much more pronounced. This binary behavior is typical for SMA actuators\cite{hong2015experimental,zhao2023starblocks}. We also noticed that the stiffness increase did not reach a plateau when the current became 0.4 A. However, we did not increase the current beyond 0.4 A since higher current can make SMA unstable and fatigue quickly over time\cite{huang2018chasing}.

\section{Design and Fabrication of the TripodBot}
\label{sec:Design&Fabrication}
We designed a tripod-shaped robot with three CCPJ straight beams to exemplify our CCPJ-based robots. This configuration is simple to fabricate and control. It also provides potential for omnidirectional locomotion for future development for field applications. 

\begin{figure*}[t]
  \centering
  \includegraphics[trim=0in 12.2cm 0in 0in, clip=true, width=1\textwidth]{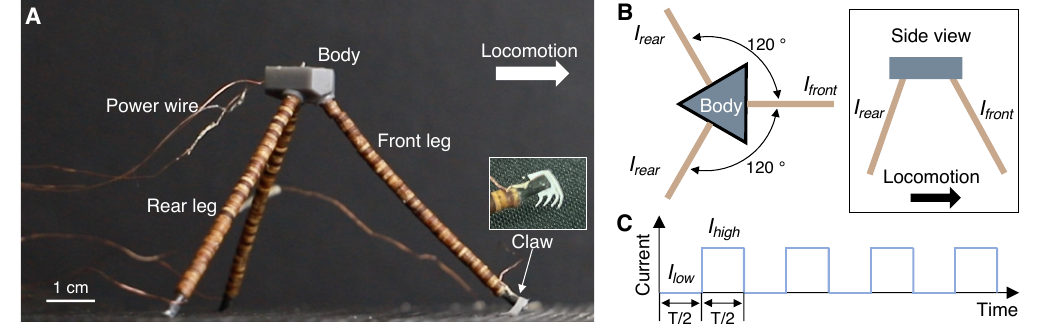}
  \caption{\textbf{Design and control of the TripodBot.} (A) Picture of the TripodBot with labels. (B) The top view and side view of the robot. (C) Typical control signal: square wave with a duty cycle of 1/2.}
  \label{fig:Design_control_TripodBot}
\end{figure*}

\subsection{Design and Control for Locomotion}
The TripodBot consists of three legs that are connected to a central body (Fig.\ref{fig:Design_control_TripodBot}A). The front leg has a specially designed claw that has anisotropic friction between the forward and backward movements. When sliding forward, the claw generates negligible friction, while preferably providing infinite friction when moving backward. The two rear legs have relatively pointy tips that also can provide anisotropic friction when interacting with the ground. The three legs are evenly distributed around the body, with the front leg pointing to the direction of locomotion (Fig.\ref{fig:Design_control_TripodBot}B). To make the control for locomotion simple, we used a square wave with a duty cycle of 1/2 (Fig.\ref{fig:Design_control_TripodBot}C). We grouped three legs into two subgroups. Two rear legs are grouped together, while the front leg is controlled individually. This design is to enable different locomotion gaits while remaining relatively simple. 

\subsection{Fabrication and Assembly}
Three identical straight legs of 65 mm long are attached to the body, with the tiled angles against the ground as 60$^\circ$. The current is applied across the SMA wires through thin copper wires, which reduces their drag on the locomotion of the robot. An origami claw was laser cut and attached on the low end of the front leg to increase the asymmetricity of friction for better locomotion\cite{yan2022crawling}. Similarly, the tips of the rear legs were sharpened to improve the performance.  

\begin{figure*}[h]
  \centering
  \includegraphics[trim=0in 9.3cm 3.6in 0in, clip=true, width=0.6\textwidth]{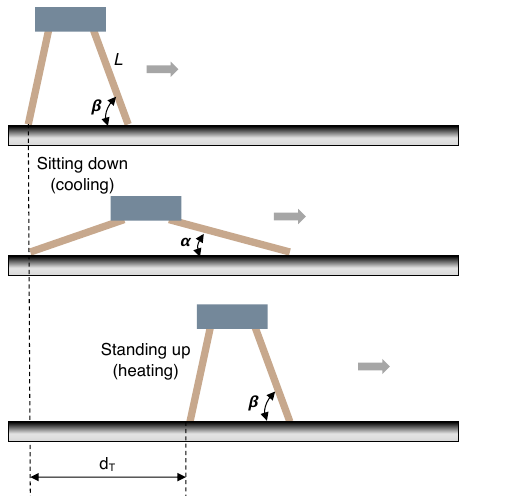}
  \caption{\textbf{Locomotion mechanism of the TripodBot.}}
  \label{fig:Model}
\end{figure*}

\subsection{Simplified Analytical Model}

To better understand the mobility of the robot, we built a simplified analytical model of its motion (Fig.\ref{fig:Model}). The robot has three legs with length \(L\). We assume the robot performs perfect stick-slip crawling on the ratchet substrate with two strokes. In the first stroke, all the three legs are softened simultaneously. Both of the rear legs hold onto anchor positions without slipping backward, while the body and front legs move forward. In the second stroke, all three legs are tightened, and the robot stands up again. We assume that the front leg can firmly grasp the ground without slipping backward, and thus the body and rear legs are pulled forward. The contact angle of the front leg with the ground is $\beta$ when standing up and is $\alpha$ when sitting down. In addition, we assume that all legs remain straight all the time. During the sitting down process, the robot moves for a distance of 
\begin{equation}
   d_{d} = \frac{L(cos\alpha-cos\beta)}{2}
\end{equation}

\noindent while it can move forward with a distance of $d_{u}$ when stands up again
\begin{equation}
   d_{u} = L(cos\alpha-cos\beta)
\end{equation}

Therefore, the speed of the robot is 
\begin{equation}
   v = \frac{d_T}{T} = \frac{d_d + d_u}{T} = \frac{3L(cos\alpha-cos\beta)}{2T}
   \label{eq:model}
\end{equation} 

The contact angles, $\alpha$ and $\beta$, are dominantly determined by a complex interaction between the legs, gravity, and ground. Because of the design, the maximum contact angle of $\beta$ for TripodBot is 60$^{\circ}$ and $\alpha$ can vary from 0 to 60$^{\circ}$.

\begin{figure*}[h]
  \centering
  \includegraphics[trim=0in 15.1cm 3.8in 0in, clip=true, width=0.6\textwidth]{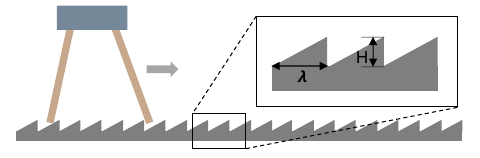}
  \caption{\textbf{Locomotion of the TripodBot on a ratchet surface.}}
  \label{fig:ratchet}
\end{figure*}

\section{Robot Demonstrations}
\label{sec:Demo}
In this section, we demonstrate the functionality---locomotion, slope climbing, adaptability, and load carrying---of the TripodBot. For simplicity and better locomotion speed, we choose to have the $I_{high}$ and $I_{low}$ of the control current as 0.4 A and 0 A according to the result from the characterization of a CCPJ beam. 

\subsection{Self-Deployment and Self-Retraction}
As shown in Fig.\ref{fig:teaser}A, the robot is initially soft and can be stored compactly. After supplying the current, the robot can self-deploy into the designed tripod shape. The geometric bounding box of the robot changes from $15 \times 17 \times 73$ mm$^3$ (minimal) to $105 \times 120 \times 64 $ mm$^3$  (about 43 times larger). After the task, e.g., locomotion, the robot can collapse back to the soft state for easy and efficient storage and transportation.

\subsection{Directional Crawling}
\begin{figure*}[h]
  \centering
  \includegraphics[trim=0in 12.5cm 0in 0in, clip=true, width=1\textwidth]{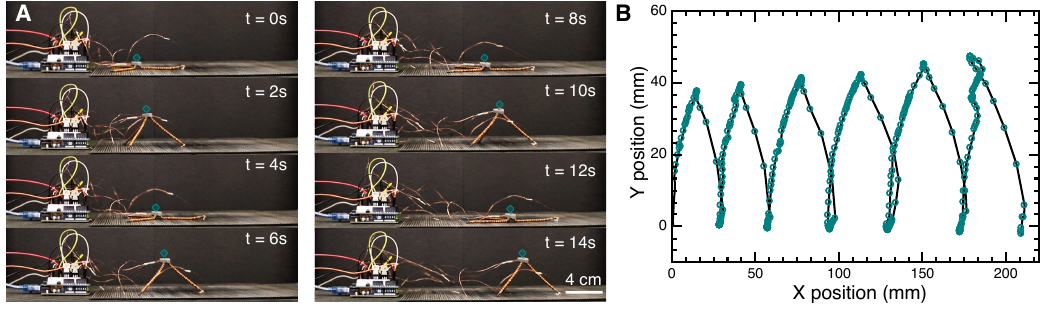}
  \caption{\textbf{Locomotion of the TripodBot with an actuation period of 4 s and applied current of 0.4A.} (A) Snapshoots. (B) Trajectory of the robot.}
  \label{fig:Locomotion}
\end{figure*}

\begin{figure*}[ht]
  \centering
  \includegraphics[trim=0in 11.7cm 0in 0in, clip=true, width=0.9\textwidth]{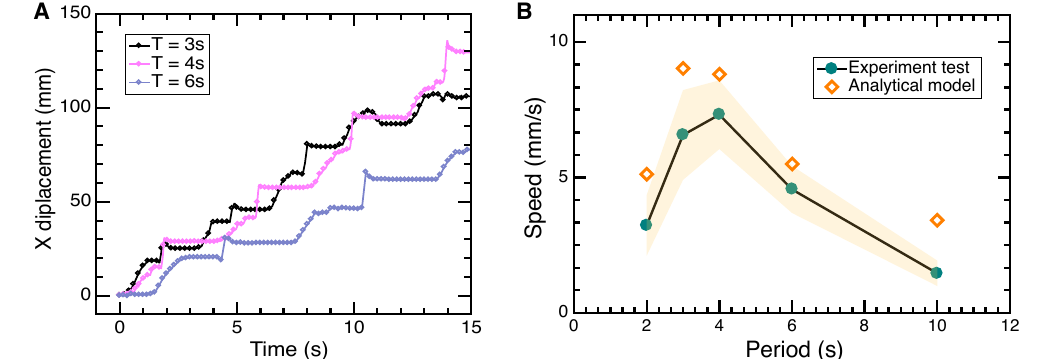}
  \caption{\textbf{Locomotion performance of the robot over actuation period.} (A) X displacement as a function of time over different actuation periods. (B) Speed over actuation period.}
  \label{fig:speed_period}
\end{figure*}

Through the stick-slip mechanism, the TripodBot can directionally crawl on the ground. We fabricated a ratchet surface ($\lambda$: 3.0 mm, $H$: 0.5 mm, see Fig. \ref{fig:ratchet}) to facilitate the locomotion. We took videos and analyzed them with the video analyzer tool \textit{Tracker} to plot the displacement and calculate average speed. When the actuation period, $T$, is 4.0 s, the robot can crawl across a distance of 209 mm with an average speed of 8.5 mm/s or 7.5 BL/min (Fig.\ref{fig:Locomotion}). By varying the actuation period from 2 s to 10 s, we showed that the robot have different locomotion speeds, reaching a maximum value at about $T = 4 $ s (Fig.\ref{fig:speed_period}).

We compared analytical and experimental results, which were all in good agreement (Fig.\ref{fig:speed_period}B). The values of $\beta$ were measured directly from tests for the simplified analytical model. Since the robot sits down fully to the ground with the selected control signal. The contact angle $\alpha$ is 0. Then according to Eq.\ref{eq:model}, the speed of the robot is
\begin{equation}
   v = \frac{3L(1-cos\beta)}{2T}
\end{equation}

We observed that the increase in $\beta$ became significantly less pronounced and approached a plateau when the actuation period exceeded 4 s. Consequently, extending the actuation period results in a reduction of the average speed. Conversely, when the period decreased below 2s, the SMA wires did not have sufficient time to heat up and/or cool down, thereby failing to generate notable locomotion. Additionally, the values predicted by the analytical model should be considered as upper bounds, as real-world tests consistently exhibit some degree of slippage.

\subsection{Slope Climbing}
The TripodBot is capable of crawling on horizontal planes as well as locomoting on inclined surfaces, thereby expanding its range of applicable environments. We demonstrated this capability by showing it climbing a slope with a 15$^{\circ}$ incline. The robot can walk across a distance of 144 mm in 60 s with an average speed of about 2.4 mm/s. This ability shows the versatility of TripodBod for real-life deployment in various terrains.

\begin{figure*}[t]
  \centering
  \includegraphics[trim=0in 14.9cm 0in 0in, clip=true, width=1\textwidth]{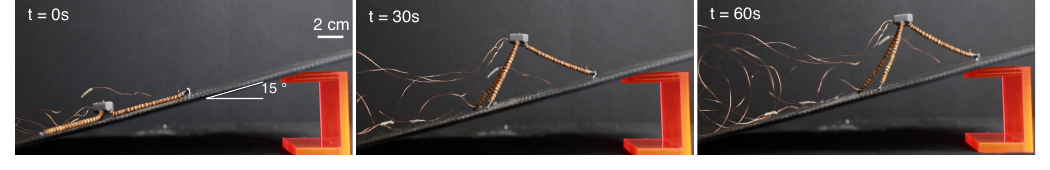}
  \caption{\textbf{Slope climbing.} The average speed is 2.4 mm/s for a slope angle of 15$^{\circ}$.}
  \label{fig:slope}
\end{figure*}

\begin{figure*}[h]
  \centering
  \includegraphics[trim=0.7in 13.3cm 0.7in 0in, clip=true, width=1\textwidth]{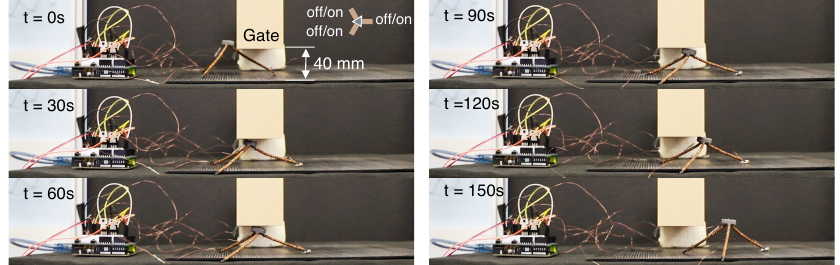}
  \caption{\textbf{The robot navigating through a confined space with the gap limit of 40 mm}. The gap is 63\% of the robot's height (63.5 mm).}
  \label{fig:Adaptivity_40}
\end{figure*}

\subsection{Adaptive Navigation in Restricted Spaces} 
Owing to the inherent softness and shape-morphing ability of the TripodBot, it can adapt its shape to the restricted environment, albeit with a reduction in its speed. We showed this adaptability by illustrating the capability to navigate through a series of restricted spaces. We first showed the robot can crawl at a relatively high speed in an open area and then lower its height to accommodate a 40 mm (63\% height) gate by reducing the applied current to 0.38 A (Fig. \ref{fig:Adaptivity_40}).

\begin{figure*}[h]
  \centering
  \includegraphics[trim=0.5in 13.1cm 0.5in 0in, clip=true, width=1\textwidth]{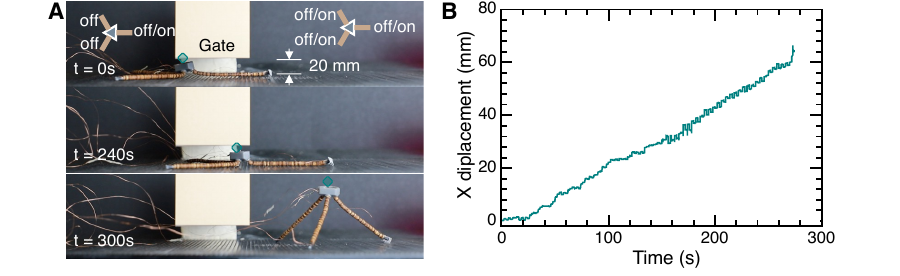}
  \caption{\textbf{The robot navigating through a confined space with the gap limit of 20 mm}. (A) The snapshoots. (B) The displacement over time. The gap is 31\% of the robot's height (63.5 mm). Only the front leg was actuated within the space.}
  \label{fig:Adaptivity_20}
\end{figure*}

Then, we showed that the robot can locomote through an even lower gate of 20 mm height (31\% height, Fig. \ref{fig:Adaptivity_20}). However, we needed to modify the control signal to only activate the front leg. Lastly, we showed that the TripodBot can navigate a narrow tunnel with a cross-section area of $40 \times 20$ mm (61\% width and 31\% height of the fully deployed TripodBot) again only activating the front leg. This capability is among the best performance of similar robots \cite{lathrop2023directionally}. 

\begin{figure*}[h]
  \centering
  \includegraphics[trim=0in 13cm 0in 0in, clip=true, width=1\textwidth]{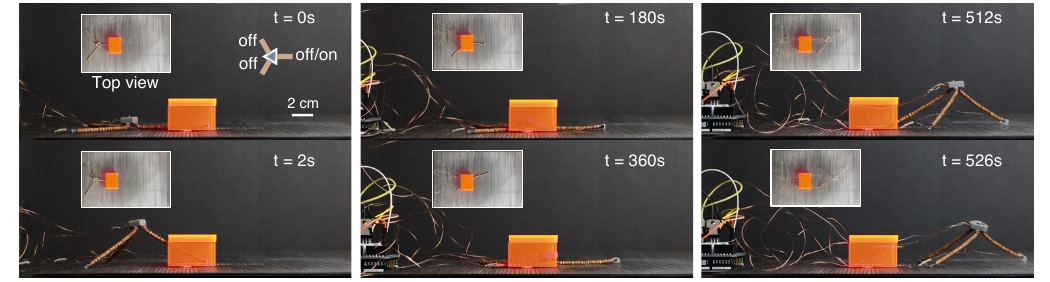}
  \caption{\textbf{The robot navigating through a confined tunnel with gap limit of 40 x 20 mm.} The robot's height and width are 63.5 mm and 66 mm, respectively. Only the front leg was actuated within the tunnel.}
  \label{fig:Adaptivity_tunnel}
\end{figure*}

\subsection{Load Carrying}
\begin{figure*}[h]
  \centering
  \includegraphics[trim=0in 13.5cm 5in 0in, clip=true, width=0.5\textwidth]{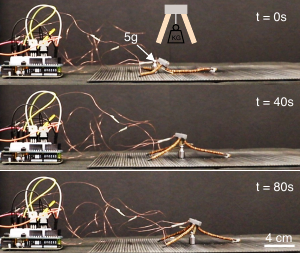}
  \caption{\textbf{Load carrying.} The average speed is 0.34 mm/s with a load of 5 g (2.4 times its weight).}
  \label{fig:load}
\end{figure*}

Loading capability is of significance for locomotive robots. The TripodBot is capable of locomotion while carrying a 5 g (2.4 times its weight) mass, as shown in Fig. \ref{fig:load}. This capability indicates the potential to integrate on-board power, control, and sensors to allow untethered operation in restricted areas. In addition, the robot can statically sustain load up to 10 g (4.8 times of its weight) with all three legs supplied with 0.4 A current. This stiffening shows potential for energy absorption for various applications, e.g., temporary protection from falling objects \cite{yan2024self,narang2017transforming}.

\subsection{Robustness and Load Bearing}
\begin{figure*}[ht]
  \centering
  \includegraphics[trim=0in 14.3cm 0in 0in, clip=true, width=1\textwidth]{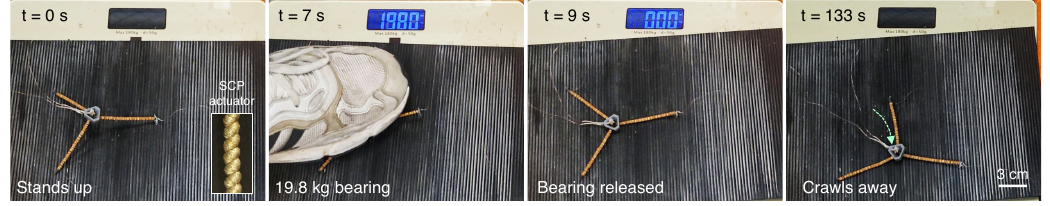}
  \caption{\textbf{Weight-bearing capability.} The robot's weight is 2.1 g. It can sustain an adult footstep of 19.8 kg force, a load about 9429 times its own body weight.}
  \label{fig:robustness}
\end{figure*}

Robustness is essential for the survival of robots in complex environments \cite{wu2019insect}. The TripodBot presented here also has exceptional robustness characteristics, resulting from the assembly of soft materials with simple structures. The robot could continue to function after being stepped on by an adult human (19.8 kg), a load about 9429 times its own body weight (Fig.\ref{fig:robustness}). Note that we used SCP actuators instead of SMA wires to improve the robustness. 

\section{Conclusions}
\label{sec:Conclusions}
% need to add "and SCP actuators" right after SMA
In this paper, we have introduced a new class of soft robots that leverage contracting-cord particle jamming (CCPJ) to achieve self-assembly, shape morphing, and stiffness modulation within a compact, integrated design. Our primary demonstration, TripodBot, exemplifies the capabilities of these robots, showcasing their potential for applications in complex and remote environments.

The development of CCPJ-based soft robots represents a significant advancement in the field of soft robotics. These robots offer numerous advantages, including ease of fabrication, low cost, higher scalability, and robustness to punctures. Additionally, they do not require bulky components like pumps and valves, making them ideal for untethered operation in challenging environments.

Our work provides a practical framework for designing soft robots with enhanced adaptability and functionality. The integration of self-assembly, shape morphing, and stiffness modulation within a single actuation system reduces implementation complexity and weight, making these robots suitable for a wide range of applications, from exploration in remote and hazardous areas to search and rescue missions in confined spaces.

Several future directions aim to enhance TripodBot and more general CCPJ-based robots: (1) Investigating new leg designs to eliminate reliance on ratchet surfaces for locomotion, making the robot deployable in real-life scenarios and unlocking omnidirectional movement due to its symmetric configuration. (2) Developing new locomotion mechanisms to navigate uneven surfaces, increasing versatility across various terrains. (3) Exploring alternative actuation options, such as electromagnetic motors, to improve actuation frequency and energy efficiency beyond the current thermal actuators (SMAs and SCP actuators), especially for larger-scale robots. (4) Creating a comprehensive analytical model to optimize performance and make robot creation more accessible to the public. (5) Current CCPJ-based robots can only have single, predefined shape morphing. We will explore new strategies to achieve multiple shape self-deployment\cite{demaine2024graph}. (6) Integrating onboard sensing, control, and power systems to enable untethered operation for field applications. By continuing to draw inspiration from biological systems and leveraging advancements in material science and robotics, we aim to develop even more capable and adaptable soft robotic systems.

In conclusion, embedding mechanical intelligence into robot bodies through the incorporation of CCPJ introduces self-deployability, stiffness tuning, and shape morphing, marking a promising step forward in the development of adaptive, self-deployable robotic systems. TripodBot serves as a compelling example of the potential of these technologies, demonstrating their ability to navigate and perform tasks in environments challenging for traditional robots. The continued exploration and development of CCPJ mechanisms will undoubtedly open new avenues for innovation and application in the field of soft robotics, contributing significantly to the advancement of autonomous robotic solutions in diverse and demanding settings.

\bibliographystyle{unsrt}
\bibliography{reference}

\begin{thebibliography}{10}

\bibitem{chen2024scale}
Tianyu Chen, Xudong Yang, Bojian Zhang, Junwei Li, Jie Pan, and Yifan Wang.
\newblock Scale-inspired programmable robotic structures with concurrent shape
  morphing and stiffness variation.
\newblock {\em Science Robotics}, 9(92):eadl0307, 2024.

\bibitem{felton2014method}
Samuel Felton, Michael Tolley, Erik Demaine, Daniela Rus, and Robert Wood.
\newblock A method for building self-folding machines.
\newblock {\em Science}, 345(6197):644--646, 2014.

\bibitem{yan2024self}
Wenzhong Yan, Talmage Jones, Christopher~L Jawetz, Ryan~H Lee, Jonathan~Brigham
  Hopkins, and Ankur Mehta.
\newblock Self-deployable contracting-cord metamaterials with tunable
  mechanical properties.
\newblock {\em Materials Horizons}, 2024.

\bibitem{lathrop2023directionally}
Emily Lathrop, Michael~T Tolley, and Nick Gravish.
\newblock Directionally compliant legs enabling crevasse traversal in small
  ground-based robots.
\newblock {\em Advanced Intelligent Systems}, 5(4):2200258, 2023.

\bibitem{jayaram2016cockroaches}
Kaushik Jayaram and Robert~J Full.
\newblock Cockroaches traverse crevices, crawl rapidly in confined spaces, and
  inspire a soft, legged robot.
\newblock {\em Proceedings of the National Academy of Sciences},
  113(8):E950--E957, 2016.

\bibitem{sun2023embedded}
Jiefeng Sun, Elisha Lerner, Brandon Tighe, Clint Middlemist, and Jianguo Zhao.
\newblock Embedded shape morphing for morphologically adaptive robots.
\newblock {\em Nature Communications}, 14(1):6023, 2023.

\bibitem{stilli2014shrinkable}
Agostino Stilli, Helge~A Wurdemann, and Kaspar Althoefer.
\newblock Shrinkable, stiffness-controllable soft manipulator based on a
  bio-inspired antagonistic actuation principle.
\newblock In {\em 2014 IEEE/RSJ International Conference on Intelligent Robots
  and Systems}, pages 2476--2481. IEEE, 2014.

\bibitem{babu2019antagonistic}
SP~Murali Babu, Ali Sadeghi, Alessio Mondini, and Barbara Mazzolai.
\newblock Antagonistic pneumatic actuators with variable stiffness for soft
  robotic applications.
\newblock In {\em 2019 2nd IEEE International Conference on Soft Robotics
  (RoboSoft)}, pages 283--288. IEEE, 2019.

\bibitem{laschi2016soft}
Cecilia Laschi, Barbara Mazzolai, and Matteo Cianchetti.
\newblock Soft robotics: Technologies and systems pushing the boundaries of
  robot abilities.
\newblock {\em Science robotics}, 1(1):eaah3690, 2016.

\bibitem{hwang2022shape}
Dohgyu Hwang, Edward~J Barron~III, ABM~Tahidul Haque, and Michael~D Bartlett.
\newblock Shape morphing mechanical metamaterials through reversible
  plasticity.
\newblock {\em Science robotics}, 7(63):eabg2171, 2022.

\bibitem{majidi2010tunable}
Carmel Majidi and Robert~J Wood.
\newblock Tunable elastic stiffness with microconfined magnetorheological
  domains at low magnetic field.
\newblock {\em Applied Physics Letters}, 97(16), 2010.

\bibitem{althoefer2018antagonistic}
Kaspar Althoefer.
\newblock Antagonistic actuation and stiffness control in soft inflatable
  robots.
\newblock {\em Nature Reviews Materials}, 3(6):76--77, 2018.

\bibitem{yang2021reprogrammable}
Bilige Yang, Robert Baines, Dylan Shah, Sreekalyan Patiballa, Eugene Thomas,
  Madhusudhan Venkadesan, and Rebecca Kramer-Bottiglio.
\newblock Reprogrammable soft actuation and shape-shifting via tensile jamming.
\newblock {\em Science Advances}, 7(40):eabh2073, 2021.

\bibitem{wall2015selective}
Vincent Wall, Raphael Deimel, and Oliver Brock.
\newblock Selective stiffening of soft actuators based on jamming.
\newblock In {\em 2015 IEEE International Conference on Robotics and Automation
  (ICRA)}, pages 252--257. IEEE, 2015.

\bibitem{narang2018mechanically}
Yashraj~S Narang, Joost~J Vlassak, and Robert~D Howe.
\newblock Mechanically versatile soft machines through laminar jamming.
\newblock {\em Advanced Functional Materials}, 28(17):1707136, 2018.

\bibitem{hasegawa2023tension}
Daniel Hasegawa, Buse Akta{\c{s}}, and Robert~D Howe.
\newblock Tension jamming for deployable structures.
\newblock In {\em 2023 IEEE/RSJ International Conference on Intelligent Robots
  and Systems (IROS)}, pages 446--451. IEEE, 2023.

\bibitem{amend2012positive}
John~R Amend, Eric Brown, Nicholas Rodenberg, Heinrich~M Jaeger, and Hod
  Lipson.
\newblock A positive pressure universal gripper based on the jamming of
  granular material.
\newblock {\em T-RO}, 28(2):341--350, 2012.

\bibitem{yan2023towards}
Wenzhong Yan and Ankur Mehta.
\newblock Towards one-dollar robots: An integrated design and fabrication
  strategy for electromechanical systems.
\newblock {\em Robotica}, 41(1):31--47, 2023.

\bibitem{huang2018chasing}
Xiaonan Huang, Kitty Kumar, Mohammad~K Jawed, Amir~M Nasab, Zisheng Ye,
  Wanliang Shan, and Carmel Majidi.
\newblock Chasing biomimetic locomotion speeds: Creating untethered soft robots
  with shape memory alloy actuators.
\newblock {\em Science Robotics}, 3(25):eaau7557, 2018.

\bibitem{hong2015experimental}
Jie Hong, Wenzhong Yan, Yanhong Ma, Dayi Zhang, and Xin Yang.
\newblock Experimental investigation on the vibration tuning of a shell with a
  shape memory alloy ring.
\newblock {\em Smart Mater. Struct.}, 24(10):105007, 2015.

\bibitem{zhao2023starblocks}
Luyang Zhao, Yijia Wu, Wenzhong Yan, Weishu Zhan, Xiaonan Huang, Joran Booth,
  Ankur Mehta, Kostas Bekris, Rebecca Kramer-Bottiglio, and Devin Balkcom.
\newblock Starblocks: Soft actuated self-connecting blocks for building
  deformable lattice structures.
\newblock {\em IEEE Robotics and Automation Letters}, 8(8):4521--4528, 2023.

\bibitem{yan2022crawling}
Wenzhong Yan and Ankur Mehta.
\newblock A crawling robot driven by a folded self-sustained oscillator.
\newblock In {\em 2022 IEEE 5th International Conference on Soft Robotics
  (RoboSoft)}, pages 455--460. IEEE, 2022.

\bibitem{narang2017transforming}
Yashraj~S Narang, Alperen Degirmenci, Joost~J Vlassak, and Robert~D Howe.
\newblock Transforming the dynamic response of robotic structures and systems
  through laminar jamming.
\newblock {\em IEEE Robotics and Automation Letters}, 3(2):688--695, 2017.

\bibitem{wu2019insect}
Yichuan Wu, Justin~K Yim, Jiaming Liang, Zhichun Shao, Mingjing Qi, Junwen
  Zhong, Zihao Luo, Xiaojun Yan, Min Zhang, Xiaohao Wang, et~al.
\newblock Insect-scale fast moving and ultrarobust soft robot.
\newblock {\em Science robotics}, 4(32):eaax1594, 2019.

\bibitem{demaine2024graph}
Rebecca Lin, Wenzhong Yan, Ankur Mehta, and Erik~D Demaine.
\newblock Routing reconfigurations.
\newblock In {\em ACM Symposium on Computational Fabrication}, SCF Adjunct '24,
  New York, NY, USA, 2024. Association for Computing Machinery.

\end{thebibliography}

\end{document}